\pdfoutput=1
\documentclass[11pt]{article}
\usepackage{acl}

\usepackage{times}
\usepackage{latexsym}
\usepackage{spverbatim}
\usepackage{float}
\usepackage[T1]{fontenc}
\usepackage[utf8]{inputenc}
\usepackage{microtype}
\usepackage{hyperref}
\usepackage{url}
\usepackage{graphicx}
\usepackage{multirow}
\usepackage{pgfplots} 
\pgfplotsset{compat=1.17} 
\usepackage{algpseudocode}
\usepackage{algorithm}
\usepackage{makecell}
\usepackage{booktabs}
\usepackage{threeparttable}

\title{CCQA: A New Web-Scale Question Answering Dataset for Model Pre-Training}
\author{Patrick Huber$^\dagger$\thanks{~~Work done at Meta.} , Armen Aghajanyan$^\ddagger$, Barlas O\u{g}uz$^\ddagger$ \\ \bf Dmytro Okhonko$^\ddagger$, \bf Wen-tau Yih$^\ddagger$, \bf Sonal Gupta$^\ddagger$ \and  \bf Xilun Chen$^\ddagger$ \\
$^\dagger$University of British Columbia; $^\ddagger$Meta AI\\
\scalebox{0.85}{
\texttt{huberpat@cs.ubc.ca}, \texttt{\{armenag,barlaso,oxo,scottyih,sonalgupta,xilun\}@fb.com}}
}

\begin{document}

\maketitle

\begin{abstract}
We propose a novel open-domain question-answering dataset based on the Common Crawl project. With a previously unseen number of around $130$ million multilingual question-answer pairs (including about $60$ million English data-points), we use our large-scale, natural, diverse and high-quality corpus to in-domain pre-train popular language models for the task of question-answering. In our experiments, we find that our Common Crawl Question Answering dataset (CCQA) achieves promising results in zero-shot, low resource and fine-tuned settings across multiple tasks, models and benchmarks\footnote{Our dataset generation script and CCQA pre-trained checkpoints can be found at \url{https://github.com/facebookresearch/CCQA}}.
\end{abstract}

\section{Introduction}
Open-domain question-answering (ODQA) has evolved into a core problem in Natural Language Processing (NLP), receiving growing interest from the research community \cite{raffel2020exploring, roberts2020much}. Despite the notoriously difficult challenge to correctly answer open-domain questions on arbitrary topics, recent advances of pre-trained language models (such as BERT \citep{devlin-etal-2019-bert}, BART \citep{lewis2020bart} and T5 \citep{raffel2020exploring}) have stimulated new research into additional, task-dependent pre-training steps. Specifically, recent publications show that in-domain pre-training regimes can improve models for several downstream tasks \citep{gururangan2020don}.
For open-domain question-answering, newly proposed pre-training tasks such as the Inverse Cloze Task (ICT) \citep{lee2019latent}, Body First Selection (BFS), Wiki Link Prediction (WLP) \citep{Chang2020Pre-training} and  Question Answering Infused Pre-training (QUIP) \citep{jia2021question} show consistent improvements over baselines. However, most of these approaches still rely on either unlabeled text, or synthetically generated question-answer (QA) pairs. In this paper, we explore a second, somewhat orthogonal dimension to these lines of work, examining if a web-scale collection of natural QA pairs can support ODQA through in-domain pre-training.

Per definition, an ODQA system should be able to answer any question from an arbitrary domain. We believe that to approach this ability with in-domain pre-training, a suitable dataset should address the following 5 challenges: (1) Size; ODQA requires knowledge of a wide variety of topics. The underlying dataset used for in-domain pre-training hence needs to cover this abundance of domains, requiring a web-scale dataset. (2) Naturalness; While synthetic corpora can potentially capture a wide variety of language phenomena, to understand and generate truly natural language in all facets, synthetic datasets are not sufficient. (3) Quality; Given the requirement for a diverse, large-scale dataset, high data quality in terms of cleanliness and sensibility becomes a major challenge. Given that web-scale data sources require highly automated approaches operating on noisy data, assuring data quality is non-trivial. (4) Diversity; Besides size, another challenge for any ODQA in-domain pre-training dataset is the generality of the corpus. The dataset needs to support answering many diverse questions to allow models to learn general concepts. (5) Evaluation Fairness; 
A web-scale question-answering dataset potentially overlaps with existing benchmark corpora, leading to inflated performance measures and impeding the evaluation fairness \citep{lewis2021question}.

To overcome these challenges, we propose a new large-scale dataset for open-domain question-answering called the \textbf{C}ommon \textbf{C}rawl \textbf{Q}uestion \textbf{A}nswering (CCQA) dataset.
Similar to popular datasets, such as C4 \citep{raffel2020exploring}, CCNet \citep{wenzek2020ccnet}, CC-100 \citep{conneau2020unsupervised}, HTLM \citep{aghajanyan2022htlm}, and CM3 \cite{aghajanyan2022cm3} we generate a large-scale, diverse and high-quality question-answering dataset from Common Crawl.

More specifically, Common Crawl allows us to obtain a large number of truly natural question-answer pairs, asked and answered by real humans on the web, rather than inferred through computational methods. Using the abundantly available \textit{schema.org} question annotation\footnote{\url{https://schema.org/Question}}, we generate question-answer pairs from explicit annotations, instead of heuristic rules, leading to high-quality data points.

In a large set of evaluations, we show that in-domain pre-training on our CCQA dataset achieves promising results across different settings, models and benchmarks. Using the rich information available on the web, we augment our dataset with additional data attributes beyond just question-answer pairs, such as votes, multiple (competing) answers, question summaries and intra-textual HTML markup, which can be used for a variety of tasks beyond question-answering in future work. 
Furthermore, we evaluate the diversity and evaluation fairness of our dataset by computing topic distributions and train-test overlaps with benchmark datasets, providing additional rationale regarding the quality of our data and experiments. To summarize, our main contributions in this paper are as follows:
\begin{itemize}
    \item We generate the first truly large-scale, natural question-answering dataset, containing around $130$ million unfiltered question-answer pairs ($55$M unique pairs), including about $60$ million English data points ($24$M unique pairs).
    \item We present key dataset statistics, confirming the high quality of our question-answer pairs, the wide range of diverse topics and a low overlap with existing benchmarks.
    \item We show the effectiveness of the dataset for in-domain pre-training by evaluating the performance of the unfiltered English subset on two question-answering tasks, three different settings, four models and five diverse benchmarks. 
\end{itemize}

\begin{figure}
    \centering
    \includegraphics[width=.85\linewidth]{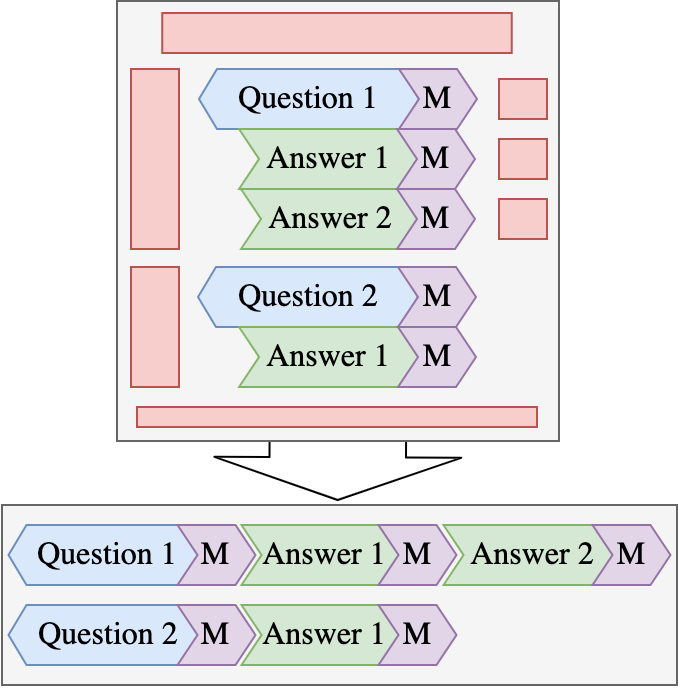}
    \caption{Dataset generation overview from the initial raw HTML file (top) to general purpose, webpage aggregated question-answer pairs (bottom). M = Additional question/answer metadata. Red boxes = Non-question-answer related webpage components.}
    \label{fig:dataset}
\end{figure}

\section{Related Work}
This work is inspired by a range of previous approaches using Common Crawl web-data, such as the Colossal Clean Crawled Corpus (C4) for language model pre-training \citep{raffel2020exploring}, the word/sentence representation generation corpus CCNet \citep{wenzek2020ccnet}, the CC-100 dataset for translation \citep{conneau2020unsupervised} and the markup-style language modelling HTLM corpus for zero-shot summarization \citep{aghajanyan2022htlm}. Despite all previously mentioned applications directly relying on large-scale web data from Common Crawl, their scope and application vary significantly. Compared to previously proposed datasets based on Common Crawl, we are the first to extract well-structured question-answer pairs with additional meta-data, making our corpus a valuable resource for ODQA research, and a multitude of related tasks, such as question summarization, answer rating, and answer ranking. 

Further web-based datasets outside the Common Crawl domain are the TriviaQA \cite{joshi2017triviaqa} and ELI5 corpora \cite{fan2019eli5}, extracting small-scale question-answer datasets from Trivia websites and Reddit threads respectively. The large-scale GooAQ dataset \cite{khashabi2021gooaq} is similarly based on web data, however exploits the Google auto-complete feature and related answer boxes to generate semi-synthetic question-answer pairs. As a large-scale, completely synthetic dataset, the PAQ corpus \cite{lewis2021paq} automatically generates a large set of \textit{Probably Asked Questions} from Wikipedia articles. In contrast to these previously proposed datasets, our CCQA corpus presents a large-scale, natural and diverse question-answering resource in the same order of magnitude as the largest synthetic datasets.

Besides the generation of the CCQA dataset, we evaluate its potential as an in-domain pre-training corpus for open-domain question-answering. Our work is aligned with previous in-domain pre-training approaches, which have shown to improve a variety of downstream tasks \citep{gururangan2020don}. Similar to in-domain pre-training, multiple domain-dependent pre-training tasks have been proposed for open-domain question-answering. For example, \citet{lee2019latent} propose the Inverse Cloze Task (ICT),  \citet{Chang2020Pre-training} introduce Body First Selection (BFS) and Wiki Link Prediction (WLP) and \citet{jia2021question} describe a novel Question Answering Infused Pre-training (QUIP) task. Along similar lines, \citet{aghajanyan2021muppet} propose pre-finetuning, an alternative to in-domain pre-training, using around $50$ domain-dependent datasets, showing that their MUPPET approach generalizes well
to many tasks. \citet{khashabi2020unifiedqa} introduce a similar concept for question-answering in their UnifiedQA framework. While we propose a somewhat orthogonal dimension to most of these works, they nevertheless present us with strong intuition regarding the effectiveness of domain-dependent pre-training.

\begin{figure}[t]
    \centering
    \includegraphics[width=\linewidth]{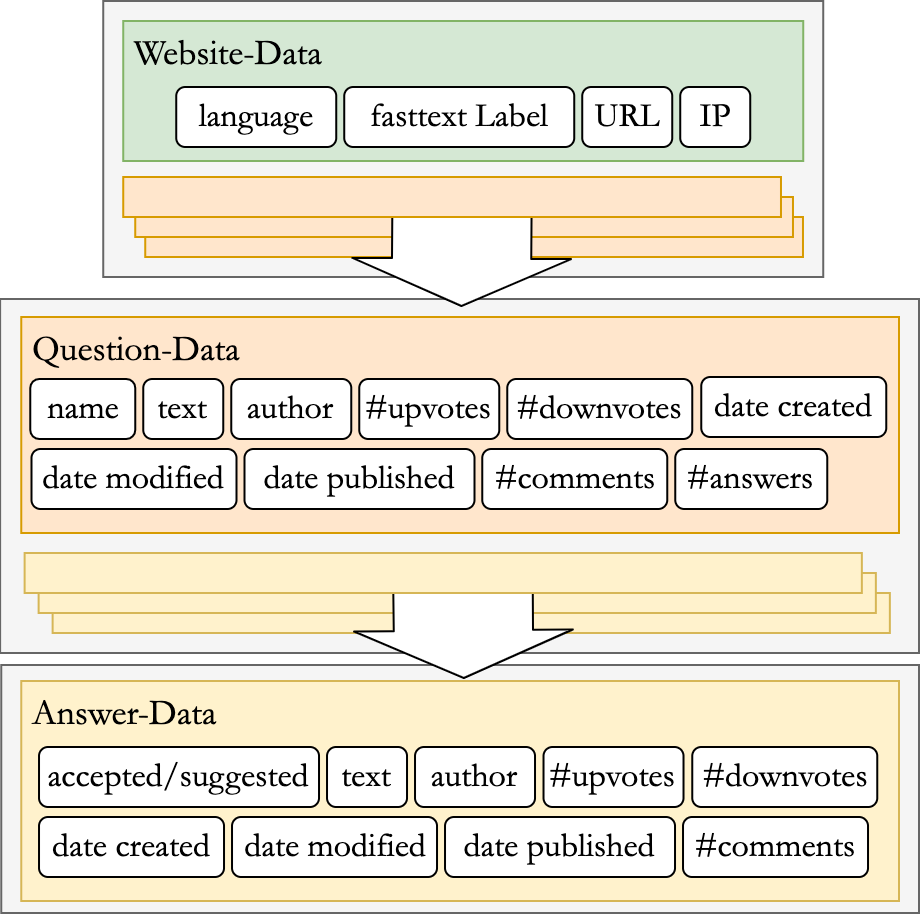}
    \caption{JSON data structure following the schema.org annotation. Fasttext language labels \citep{joulin2016fasttext, joulin2017bag} added for language distinction.}
    \label{fig:format}
\end{figure}

\section{The Common Crawl Question Answering (CCQA) Dataset}
\subsection{Dataset Collection}
\label{data_collection}
Our Common Crawl Question Answering (CCQA) dataset contains around $130$ million question-answer pairs ($55$M unique), extracted from 13 Common Crawl snapshots between May 2020 and May 2021\footnote{\url{https://commoncrawl.org/}}. A high-level overview of the dataset generation process is depicted in Figure~\ref{fig:dataset}. Starting from a set of raw HTML webpages, we make use of the schema.org definition to find relevant tags, such as the question, answer, author and votes (for the full set of tags see Figure~\ref{fig:format}). Using the explicit schema.org annotation (commonly used for search-engine optimization), instead of simple heuristics (e.g. question marks), we optimize the resulting corpus for high-quality data points. Specifically, due to the added efforts for website creators to define schema.org conforming meta-data, we believe that annotated question-answer pairs are likely to be relevant to the general public, mostly exclude rhetorical and contextual questions, and as a result constitute high quality QA data, despite the noisy nature of webpages.

During the dataset processing steps, we remove all HTML elements that do not contain valid schema.org markers (red in Figure~\ref{fig:dataset}) and subsequently clean every question on the webpage to only conserve markup related to the textual content of schema.org tags\footnote{Set of textual tags taken from \url{developer.mozilla.org/en-US/docs/Web/HTML/Element}}.
We further remove any unrelated markup attributes (e.g., CSS and JavaScript classes), before converting the content into a well-defined JSON object, shown in Figure~\ref{fig:format} and further described in section \ref{sec:format}. 

Using the 13 consecutive Common Crawl snapshots, we generate an initial dataset of $130$ million question-answer pairs, naturally containing two types of potential duplicates: (1) Same-URL duplicates; where a webpage is updated between any two Common Crawl snapshots and (2) Content duplicates; where webpages from any Common Crawl snapshot contain same questions with potentially similar answers. 

Here, we use the original, uncleaned version of the dataset, presenting a practical performance lower-bound, while leaving the exploration of additional filtering steps for future work\footnote{We provide de-duplication scripts for same-URL duplicates due to snapshot overlap at \url{https://github.com/facebookresearch/CCQA}.}.

Our dataset generation procedure is further outlined in Algorithm \ref{alg:algo_1}, found in Appendix \ref{app:algo}. For qualitative examples of our generated dataset format, we refer readers to Appendix~\ref{quali}.

\subsection{Dataset Format}
\label{sec:format}
The structured output of the dataset collection (shown in Figure~\ref{fig:format}), contains a three-level nested structure: (1) Every top-level data point represents a webpage in Common Crawl, encapsulating questions and answers found on the page, together with relevant metadata.
(2) On the second level, every question is represented as a tuple containing the question name (a short summary of the question) and question text (the main question). Questions also contain additional metadata as shown in Figure~\ref{fig:format}. 
(3) Every question can contain an arbitrary number of associated answers and answer attempts, located on the third and final level of the nested structure. An answer thereby contains a mandatory accepted/suggested label, the answer text as well as optional metadata.

With this nested structure of our CCQA dataset, we allow users to verify question-answer pairs and their metadata on the original webpage, utilize additional parts of the web-document and allow future research to tackle question-answering related tasks, such as answer selection, answer rating or answer ranking.

\subsection{Dataset Dimensions} 
\label{data_dimension}
To gain better insights into the massive amount of data, we present a mix of automatically obtained dataset dimensions, a small-scale human pilot study, and a set of key dataset distributions.

Regarding the small-scale human pilot study, we analyze a random subset of $400$ individual question-answer pairs and evaluate their sensibility and answerability. We define \textit{question sensibility} as to whether the annotator understands the questions itself, while
\textit{question answerability} refers to whether the question provides enough context for a perfect question-answering system to correctly answer the question. Furthermore, \emph{QA-sensibility} denotes if the question-answer pair makes sense\footnote{We do not check the answer for factual correctness but merely evaluate if it \emph{could} be the answer for the given question.}. We refer interested readers to Table~\ref{tab:sens_ans_detailed} in Appendix~\ref{app:sens_ans} for further explanations on sensibility/answerability.
 
\begin{table}[t]
    \centering
    \setlength{\tabcolsep}{0.28em}
    \scalebox{0.88}{
    \begin{tabular}{ccccc}
        \toprule
        Q-Sens$^H$ & Q-Ans$^H$ & QA-Sens$^H$ & Markup & Q-Summ \\
        \midrule
        96.5\% & 86\% & 82.25\% & 47.5\% & 11.7\% \\
        \toprule
        No A & Avg \#A$^*$ & Mean Q & Mean A & Lang Tags \\
        \midrule
        5.9\% & 1.41 & 43 & 57 & 77.9\% \\
        \bottomrule
    \end{tabular}
    }
    \caption{Key CCQA dataset dimensions. Q=Question, A=Answer, QA=\allowbreak Question-answer pair, Sens=\allowbreak Sensibility, Ans=\allowbreak Answerability, Lang=\allowbreak Language, Summ=\allowbreak Summarization, Mean=Average number of words, $^H$Human pilot study, $^*$Excluding questions without answers.}
    \label{tab:data_dimensions}
\end{table}

As shown in Table~\ref{tab:data_dimensions}, our CCQA corpus contains nearly exclusively sensible questions, with the vast majority of them also answerable and sensible as a pair. To complement our small-scale human annotation, we further explore key dataset dimension, including the fraction of samples with advanced markup, questions containing both, question name and question text (as defined by the schema.org annotation), the number of questions without gold-answers, average question and answer length and the number of webpages with a valid language label, all indicating that the schema.org annotation highly correlates with carefully curated webpages. 

Besides the key corpus-level statistics, we take a closer look at important dataset distributions in Table \ref{tab:data_distributions}. Specifically, we present the top 5 domains at the top of Table~\ref{tab:data_distributions}, showing the largest number of webpages originating from the \textit{stackexchange} domain, accounting for about $8\%$ of data points. Regarding the topical distribution of our dataset, we use the DMOZ/Curlie taxonomy, automatically extracting hierarchical topic information\footnote{\url{https://www.curlie.org}}. We randomly sample $1,000$ question webpages and show the top 5 topics in the second row of Table \ref{tab:data_distributions}. A more detailed topic distribution, also considering second-level assignments, can be found in Table~\ref{tab:data_distribution_detailed} in Appendix~\ref{app:topic_dist}.
Regarding the question-word distribution in our CCQA dataset, we observe that the majority of $36\%$ of question words are \textit{what} questions, followed by \textit{how}, \textit{when}, \textit{which} and \textit{where}. A full list of all 8 questions words and their relative appearance in our corpus can be found in Table \ref{tab:q_words_detailed} in Appendix \ref{app:q_dist}.
Lastly, expanding on the number of non-trivial markup tags presented in Table \ref{tab:data_dimensions}, we explore the frequency of HTML markup tags in our dataset in the last row in Table~\ref{tab:data_distributions}. For a list of the top-25 tags found in our corpus, we point interested readers to Table~\ref{tab:data_markup_detailed} in Appendix~\ref{app:html_dist}.

\begin{figure}[t]
    \centering
    \includegraphics[width=\linewidth]{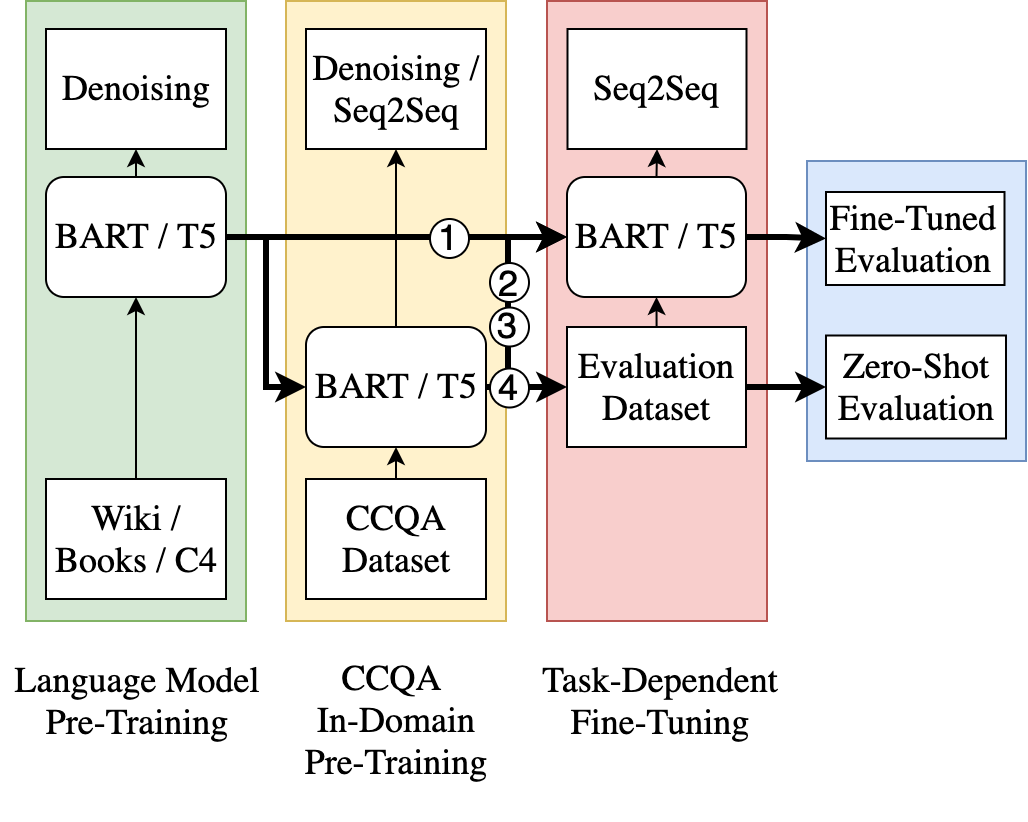}
    \caption{High-level overview of the closed-book CCQA in-domain pre-training step (yellow) as part of the training pipeline for BART and T5. Language model pre-training shown in green. Task-dependent fine-tuning presented in red. Evaluation in blue. (1) Baseline pre-training/fine-tuning pipeline, (2) In-domain pre-training/fine-tuning pipeline, (3) zero-shot baseline setting and (4) zero-shot in-domain pre-training setting.}
    \label{fig:high_level_closed_book}
\end{figure}

\begin{table*}[t]
    \centering
    \setlength{\tabcolsep}{0.2em}
    \scalebox{0.9}{
    \begin{tabular}{lrrrrr}
        \toprule
        Metric & \multicolumn{5}{c}{Top 5 Appearances in CCQA} \\
        \midrule
        Domains & stackexchange (07.78\%) & hotels (03.46\%) & viamichelin (02.51\%) & ccm (01.86\%) & vrbo (01.74\%) \\
        Topics & Regional (38.90\%) & Society (21.10\%) & Business (08.30\%) & Sports (07.00\%) & Rec (06.20\%) \\
        Q-words & What (36.20\%) & How (29.80\%) & When (09.68\%) & Which (09.64\%) & Where (06.04\%) \\
        Markup & p (28.48\%) & a (14.89\%) & br (14.86\%) & li (10.04\%)  & span (05.77\%) \\
        \bottomrule
    \end{tabular}}
    \caption{CCQA dataset distribution for top 5 domains, topics according to the DMOZ/Curlie annotation, question words (Q-words, only computed on the English subset) and most common markup tags. \% for q-words and markup tags presents portion of all q-word/markup appearances. ccm=commentcamarche, Rec=Recreational.}
    \label{tab:data_distributions}
\end{table*}

\section{Evaluation} 
In this section, we showcase the value of our CCQA dataset with experiments on the closed-book question-answering (section \ref{cbqa}) and passage retrieval  for open-book QA (section \ref{pass_ref}) tasks.

\subsection{Closed-Book Question-Answering}
\label{cbqa}
\subsubsection{Task}
The closed-book question-answering task challenges systems to answer questions without to use of additional information sources, such as knowledge bases or evidence documents. As a result, models are solely relying on the question text and the information stored inside the model weights during training. Here, we evaluate our new CCQA dataset as an in-domain pre-training corpus for this highly challenging task by converting the JSON representation into plain question-answer pairs, removing markup tags and additional metadata.

\subsubsection{Models \& Training}
Using the question-answer pairs from the CCQA dataset, we in-domain pre-train large language models for question-answering. We start with vanilla BART and T5 transformer models, shown on the left side (green) in Figure \ref{fig:high_level_closed_book}. We then further in-domain pre-train the models using a denoising or sequence-to-sequence (seq2seq) setup (yellow box in Figure~\ref{fig:high_level_closed_book}). For the denoising task, we follow the vanilla BART approach \cite{lewis2020bart}, using a concatenation of \textit{Q:$\mathbin\Vert$<question>$\mathbin\Vert$A:$\mathbin\Vert$<answer>} as the model input. For the seq2seq task, we train the model to predict the gold answer given a question as input. 
With the additional in-domain pre-training step, a variety of training-flows emerge, shown as numbered circles in Figure~\ref{fig:high_level_closed_book}: \\
(1) Using a vanilla pre-trained language model to fine-tune on the benchmark dataset. \\
(2) Using the CCQA dataset for in-domain pre-training and subsequently fine-tune on the benchmark dataset. \\
(3) Using a pre-trained language model to directly infer answers on the benchmark dataset (zero-shot). \\
(4) Using the CCQA in-domain pre-trained model to directly infer answers on the benchmark dataset in zero-shot fashion.

\subsubsection{Datasets}
\label{datasets}
We evaluate the performance of our CCQA corpus as an in-domain pre-training dataset on 5 common benchmarks, based on 4 publicly available datasets in the closed-book setting:

\textbf{TriviaQA} (TQA) is a short-form, factoid-style question-answering dataset \cite{joshi2017triviaqa}. For the closed-book task, we ignore the available contexts and focus exclusively on question-answer pairs. Since the official test-split of the dataset is not publicly available, we use the official validation set as our test split and randomly sample a validation set from the training data, as commonly done in previous work \citep{roberts2020much}.

\textbf{Natural Questions} (NQ) \citep{kwiatkowski-etal-2019-natural} represents a popular corpus for question-answering research. Despite most recent work focusing on the \textbf{short-form} answers (NQ-Short), the NQ corpus also provides additional \textbf{long-form} answers (NQ-Long) for a large subset of questions. In this work, we use both, short, factoid answers and long-form responses.

\textbf{ELI5}, introduced by \citet{fan2019eli5}, constitutes the first large-scale long-form dataset for open-ended question-answering. We again do not take available evidence documents into account, but focus on the question-answer pairs only.

\textbf{GooAQ} \citep{khashabi2021gooaq} contains semi-automatically extracted question-answer pairs from the Google question auto-complete feature.

\begin{table*}[t]
\centering
\begin{threeparttable}
\setlength{\tabcolsep}{0.3em}
\centering
\scalebox{0.95}{
\begin{tabular}{lrrrrr}
\toprule
\multicolumn{6}{c}{\bf Zero-Shot}\\
\midrule
\multirow{2}{*}{Model} & \multicolumn{1}{r}{TQA} & \multicolumn{1}{r}{NQ-Short} & \multicolumn{1}{r}{NQ-Long} & \multicolumn{1}{r}{ELI5} & \multicolumn{1}{r}{GooAQ} \\
 & \multicolumn{1}{r}{AR} & \multicolumn{1}{r}{AR} & \multicolumn{1}{r}{R-L} & \multicolumn{1}{r}{R-L} & \multicolumn{1}{r}{R-L}\\
\midrule
\multicolumn{6}{c}{BART Large}\\
\midrule
Rand. Init. & 0.04 & 0.11 & 0.10 & 0.26 & 0.16 \\
Vanilla & $^\dagger$4.91 & $^\dagger$1.93 & 10.39 & 11.88 & 14.67 \\
Vanilla$^a$ &  &  &  &  &  \\
CCQA & \bf $^\dagger$5.14 & \bf $^\dagger$2.16 & \bf 12.18 & \bf $^\dagger$15.21 & \bf $^\dagger$17.5 \\
CCQA-d & 4.80 & 2.13 & 10.33 & 11.91 & 14.88 \\
\midrule \addlinespace
\multicolumn{6}{c}{T5 Small}\\
\midrule
Rand. Init. & 0.05 & 0.11 & 1.13 & 1.49 & 0.80 \\
Vanilla & $^\dagger$5.06 & $^\dagger$1.74 & 9.16 & 7.55 & $^\dagger$8.92 \\
Vanilla$^b$ &  &  &  &  &  \\
CCQA & \bf $^\dagger$5.13 & \bf $^\dagger$1.86 & \bf $^\dagger$13.63 & \bf $^\dagger$15.28 & \bf $^\dagger$15.46 \\
\midrule \addlinespace
\multicolumn{6}{c}{T5 Base}\\
\midrule
Rand. Init. & 0.04 & 0.11 & 0.00 & 0.00 & 0.00 \\
Vanilla & $^\dagger$5.49 & $^\dagger$2.02 & $^\dagger$14.39 & 12.27 & $^\dagger$14.99 \\
Vanilla$^c$ &  &  &  &  &  \\
CCQA & \bf $^\dagger$7.15 & \bf $^\dagger$3.19 & \bf $^\dagger$15.08 & \bf $^\dagger$15.69 & \bf $^\dagger$15.85 \\
\bottomrule
\end{tabular}}
\centering
\scalebox{0.95}{
\begin{tabular}{rrrrr}
\toprule
\multicolumn{5}{c}{\bf Fine-Tuned}\\
\midrule 
\multicolumn{1}{r}{TQA} & \multicolumn{1}{r}{NQ-Short} & \multicolumn{1}{r}{NQ-Long} & \multicolumn{1}{r}{ELI5} & \multicolumn{1}{r}{GooAQ} \\
 \multicolumn{1}{r}{EM} & \multicolumn{1}{r}{EM} & \multicolumn{1}{r}{R-L} & \multicolumn{1}{r}{R-L} & \multicolumn{1}{r}{R-L}\\
\midrule 
\multicolumn{5}{c}{BART Large}\\
\midrule
 0.71 & 0.75 & 16.04 & 14.37 & 16.21 \\
 \bf 28.67 & 23.79 & 23.47 & 16.96 & 35.67 \\
  & 26.50 &  &  &  \\
 25.82 & 22.91 & 21.25 & 17.23 & 32.53 \\
 27.84 & \bf 23.96 & \bf 24.56 & \bf 17.27 & \bf 35.92 \\
\midrule \addlinespace
\multicolumn{5}{c}{T5 Small}\\
\midrule
 0.44 & 0.54 & 10.86 & 13.06 & 8.71 \\
 \bf 21.02 & \bf 21.16 & \bf 22.09 & 16.28 & 24.70 \\
  &  & 19.00 & 23.00 \\
 17.55 & 19.50 & 22.05 & \bf 16.33 & \bf 25.35 \\
\midrule \addlinespace
\multicolumn{5}{c}{T5 Base}\\
\midrule
 0.32 & 0.38 & 13.58 & 12.72 & 7.93 \\
 \bf 26.25 & \bf 23.04 & \bf 25.36 & 16.58 & \bf 29.36 \\
 23.63 & 25.94 &  &  &  \\
 22.69 & 22.32 & 24.73 & \bf 16.64 & 29.09 \\
\bottomrule
\end{tabular}
}
\begin{tablenotes}[para,raggedright]
\hbox{Results from 
\item[a] \citet{lewis2021question}
\item[b] \citet{khashabi2021gooaq}
\item[c] \citet{roberts2020much}}
\end{tablenotes}
\caption{Closed-book zero-shot and fine-tuned results. Best performance of fairly computed results per sub-table \textbf{bold}. $^\dagger$Zero-shot model outperforms fully fine-tuned randomly initialized transformer of same architecture. \textit{-d} extension indicates denoising CCQA pre-training task. \textit{AR}=Answer-level recall, \textit{EM}=Exact Match, \textit{RL}=Rouge-L.
}
\label{tab:cb_results}
\end{threeparttable}
\end{table*}

\subsubsection{Metrics} 
\label{metrics}
For datasets with short-form answers, we use the Exact Match (EM) metric for fine-tuned systems, in line with previous work by \citet{roberts2020much} and \citet{lewis2021paq}. While the EM metric works well for systems that are aware of the task-specific format, it punishes potentially correct answers with additional context, which we believe is overly harsh in zero-shot settings, where the specific output format is not known (e.g., training-flows (3) and (4)). Therefore, we propose using the Answer-level Recall (AR) metric for our zero-shot experiments, while limiting the answer length with the \textit{max-length} and \textit{length-penalty} inference parameters. As such, the AR metric requires the correct answer to be a continuous sub-sequence of the predicted tokens, while allowing for additional context. Since AR operates on token-level, the prediction of super/sub-words, e.g., \textit{\textbf{fun}damental} instead of \textit{\textbf{fun}}, is considered incorrect.

For long-form question-answer datasets, we choose the Rouge-L (RL) score as our evaluation metric, which has shown strong correlation with Rouge-1 and Rouge-2 scores, and is commonly used in previous work \cite{khashabi2021gooaq}.

\subsubsection{Hyper-Parameters}
We use the default parameters of the BART \citep{lewis2020bart} and T5 \citep{raffel2020exploring} models for in-domain pre-training and fine-tuning whenever possible. Regarding the in-domain pre-training on our CCQA dataset, we limit training to $800k$ steps using a batch-size of $1{,}024$. During our fine-tuning runs, we limit the number of updates to $20k$ steps with a batch-size of $256$ samples, with exception of the GooAQ dataset, which we fine-tune for $100k$ steps due to it's large size. We select the best model during our in-domain pre-training runs based on the perplexity measure, and pick the top fine-tuned model according to the final evaluation metric. We do not perform any hyper-parameter search during in-domain pre-training and fine-tuning.

For the inference step, our hyper-parameter setting is closely related to commonly used summarization parameters. We use a beam-size of 4, max-length of 140, and length-penalty of 2.0. For the fine-tuned short-form task, we choose a max-length of 30, following \citet{xiong2021answering} and a length-penalty of 1.0. All model evaluations are based on Huggingface Transformers\footnote{Experiments are executed on Nvidia V100 32GB GPUs.} \cite{wolf2019huggingface}.

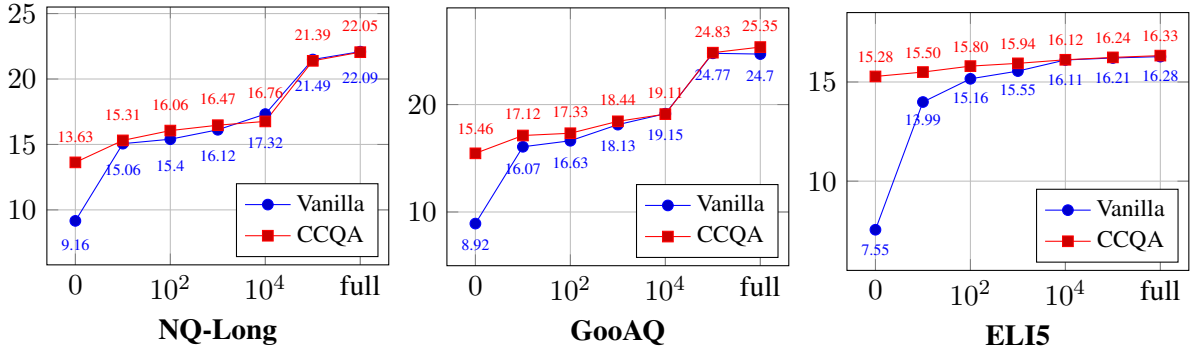
\begin{figure*}[t]
    \centering
    \begin{tikzpicture}
        \begin{axis}[
                width=0.38*\textwidth,
                height=5cm,
                legend pos=south east,
                legend style={font=\small},
                ymajorgrids=true,
                xmajorgrids=true,
                bar width=1.5mm,
                ymin=5.8, ymax=25.5,
                xlabel={\bf NQ-Long},
                symbolic x coords={$0$, $10^1$, $10^2$, $10^3$, $10^4$, $10^5$, full},
            ]
            \addplot coordinates {
                ($0$, 9.16) ($10^1$, 15.06) ($10^2$, 15.4) ($10^3$, 16.12) ($10^4$, 17.32) ($10^5$, 21.49) (full, 22.09)
            };
            \addlegendentry{Vanilla}
            \node at (axis cs:$0$,7.36){\tiny\textcolor{blue}{9.16}};
            \node at (axis cs:$10^1$,13.06){\tiny\textcolor{blue}{15.06}};
            \node at (axis cs:$10^2$,13.4){\tiny\textcolor{blue}{15.4}};
            \node at (axis cs:$10^3$,14.12){\tiny\textcolor{blue}{16.12}};
            \node at (axis cs:$10^4$,15.12){\tiny\textcolor{blue}{17.32}};
            \node at (axis cs:$10^5$,19.49){\tiny\textcolor{blue}{21.49}};
            \node at (axis cs:full,20.09){\tiny\textcolor{blue}{22.09}};
            
            \addplot coordinates {
                 ($0$, 13.63) ($10^1$, 15.31) ($10^2$, 16.06) ($10^3$, 16.47) ($10^4$, 16.76) ($10^5$, 21.39) (full, 22.05)
            };
            \addlegendentry{CCQA}
            \node at (axis cs:$0$,15.63){\tiny\textcolor{red}{13.63}};
            \node at (axis cs:$10^1$,17.31){\tiny\textcolor{red}{15.31}};
            \node at (axis cs:$10^2$,18.06){\tiny\textcolor{red}{16.06}};
            \node at (axis cs:$10^3$,18.47){\tiny\textcolor{red}{16.47}};
            \node at (axis cs:$10^4$,18.99){\tiny\textcolor{red}{16.76}};
            \node at (axis cs:$10^5$,23.39){\tiny\textcolor{red}{21.39}};
            \node at (axis cs:full,24.05){\tiny\textcolor{red}{22.05}};
            
        \end{axis}
    \end{tikzpicture}
    \begin{tikzpicture}
        \begin{axis}[
                width=0.38*\textwidth,
                height=5cm,
                legend style={font=\small},
                legend pos=south east,
                ymajorgrids=true,
                xmajorgrids=true,
                bar width=1.5mm,
                ymin=5, ymax=29,
                xlabel={\bf GooAQ},
                symbolic x coords={$0$, $10^1$, $10^2$, $10^3$, $10^4$, $10^5$, full},
            ]
            \addplot coordinates {
                ($0$, 8.92) ($10^1$, 16.07) ($10^2$, 16.63) ($10^3$, 18.13) ($10^4$, 19.15) ($10^5$, 24.77) (full, 24.7)
            };
            \addlegendentry{Vanilla}
            \node at (axis cs:$0$,6.92){\tiny\textcolor{blue}{8.92}};
            \node at (axis cs:$10^1$,14.07){\tiny\textcolor{blue}{16.07}};
            \node at (axis cs:$10^2$,14.63){\tiny\textcolor{blue}{16.63}};
            \node at (axis cs:$10^3$,16.13){\tiny\textcolor{blue}{18.13}};
            \node at (axis cs:$10^4$,17.15){\tiny\textcolor{blue}{19.15}};
            \node at (axis cs:$10^5$,22.77){\tiny\textcolor{blue}{24.77}};
            \node at (axis cs:full,22.7){\tiny\textcolor{blue}{24.7}};
            
            \addplot coordinates {
                 ($0$, 15.46) ($10^1$, 17.12) ($10^2$, 17.33) ($10^3$, 18.44) ($10^4$, 19.11) ($10^5$, 24.83) (full, 25.35)
            };
            \addlegendentry{CCQA}
            \node at (axis cs:$0$,17.76){\tiny\textcolor{red}{15.46}};
            \node at (axis cs:$10^1$,19.12){\tiny\textcolor{red}{17.12}};
            \node at (axis cs:$10^2$,19.33){\tiny\textcolor{red}{17.33}};
            \node at (axis cs:$10^3$,20.44){\tiny\textcolor{red}{18.44}};
            \node at (axis cs:$10^4$,21.11){\tiny\textcolor{red}{19.11}};
            \node at (axis cs:$10^5$,26.83){\tiny\textcolor{red}{24.83}};
            \node at (axis cs:full,27.35){\tiny\textcolor{red}{25.35}};
            
        \end{axis}
    \end{tikzpicture}
    \begin{tikzpicture}
        \begin{axis}[
                width=0.38*\textwidth,
                height=5cm,
                legend pos=south east,
                legend style={font=\small},
                ymajorgrids=true,
                xmajorgrids=true,
                bar width=1.5mm,
                ymin=5.5, ymax=18.5,
                xlabel={\bf ELI5},
                symbolic x coords={$0$, $10^1$, $10^2$, $10^3$, $10^4$, $10^5$, full},
            ]
            \addplot coordinates {
                ($0$, 7.55) ($10^1$, 13.99) ($10^2$, 15.16) ($10^3$, 15.55) ($10^4$, 16.11) ($10^5$, 16.21) (full, 16.28)
            };
            \addlegendentry{Vanilla}
            \node at (axis cs:$0$,6.55){\tiny\textcolor{blue}{7.55}};
            \node at (axis cs:$10^1$,13.1){\tiny\textcolor{blue}{13.99}};
            \node at (axis cs:$10^2$,14.2){\tiny\textcolor{blue}{15.16}};
            \node at (axis cs:$10^3$,14.6){\tiny\textcolor{blue}{15.55}};
            \node at (axis cs:$10^4$,15.1){\tiny\textcolor{blue}{16.11}};
            \node at (axis cs:$10^5$,15.2){\tiny\textcolor{blue}{16.21}};
            \node at (axis cs:full,15.3){\tiny\textcolor{blue}{16.28}};

            \addplot coordinates {
                 ($0$, 15.28) ($10^1$, 15.5) ($10^2$, 15.80) ($10^3$, 15.94) ($10^4$, 16.12) ($10^5$, 16.24) (full, 16.33)

            };
            \addlegendentry{CCQA}
            \node at (axis cs:$0$,16.28){\tiny\textcolor{red}{15.28}};
            \node at (axis cs:$10^1$,16.5){\tiny\textcolor{red}{15.50}};
            \node at (axis cs:$10^2$,16.8){\tiny\textcolor{red}{15.80}};
            \node at (axis cs:$10^3$,16.94){\tiny\textcolor{red}{15.94}};
            \node at (axis cs:$10^4$,17.12){\tiny\textcolor{red}{16.12}};
            \node at (axis cs:$10^5$,17.24){\tiny\textcolor{red}{16.24}};
            \node at (axis cs:full,17.33){\tiny\textcolor{red}{16.33}};
            
        \end{axis}
    \end{tikzpicture}
    
    \caption{Low resource experiments comparing the Rouge-L score of vanilla T5 Small with our CCQA pre-trained models on NQ-long (left), GooAQ (center) and ELI5 (right).}
    \label{fig:low_resource}
\end{figure*}

\subsubsection{Results}
\label{closed_book}
Our main results for the closed-book question-answering task are presented in Table \ref{tab:cb_results}, showing the zero-shot and fine-tuned performance of the BART Large (top), T5 Small (center) and T5 Base (bottom) models for each of the 5 evaluation datasets. Even though we present a wide variety of benchmark results, from short-form factoid questions to long-form answers, the CCQA seq2seq pre-trained model consistently outperforms all other models on the zero-shot question-answering task. Even more importantly, the additional in-domain pre-training step achieves better zero-shot performance than fully fine-tuned, randomly initialized transformer models (as extensively used prior to 2018) in almost all settings. Specifically, our model outperforms the randomly initialized transformers on all benchmarks for T5 Small and T5 Base, as well as on 4 out of 5 datasets using BART Large. 

Comparing the fully fine-tuned setting across models and datasets it becomes clear that, although oftentimes performing comparably, our CCQA seq2seq pre-trained model underperforms the vanilla models in most cases. Seq2seq in-domain pre-training on CCQA only reaches superior performance on the ELI5 dataset for all models, as well as on the GooAQ dataset for T5 Small. 
Showing that seq2seq pre-training on CCQA is effective in zero-shot scenarios, however only partially improves over baselines in the fine-tuned setting, we investigate: (1) Additional experiments using the CCQA dataset for denoising-style pre-training (\textit{-d} in Table \ref{tab:cb_results}) and (2) Evaluate additional low-resource scenarios, shown in Figure \ref{fig:low_resource}.

For our denoising-style in-domain pre-training experiment, we keep the available markup information, in line with HTLM \cite{aghajanyan2022htlm}. 
As shown in Table~\ref{tab:cb_results}, the in-domain CCQA denoising objective outperforms the vanilla BART Large model on 4 out of 5 benchmarks in the fine-tuned setting. We believe that this result, alongside the zero-shot performance of the seq2seq CCQA model, clearly shows the usefulness and generality of our CCQA corpus for closed-book open-domain question-answering.

Taking a closer look at low-resource scenarios, we evaluate the vanilla T5 Small model against our in-domain pre-trained approach using 5 proper subsets of the NQ-Long, GooAQ and ELI5 benchmark datasets, drawn at random. As presented in Figure \ref{fig:low_resource}, our CCQA model mostly outperforms the vanilla T5 Small model in low-resource scenarios with up to $10{,}000$ data points. While the performance of our CCQA model is consistently better on the ELI5 test-set, the vanilla baselines outperform our models fastest on the NQ-Long corpus. Additional low-resource experiments on T5 Base are shown in Table \ref{fig:app_low_resource}, in Appendix \ref{app_full_res}.

\begin{figure}[t]
    \centering
    \includegraphics[width=\linewidth]{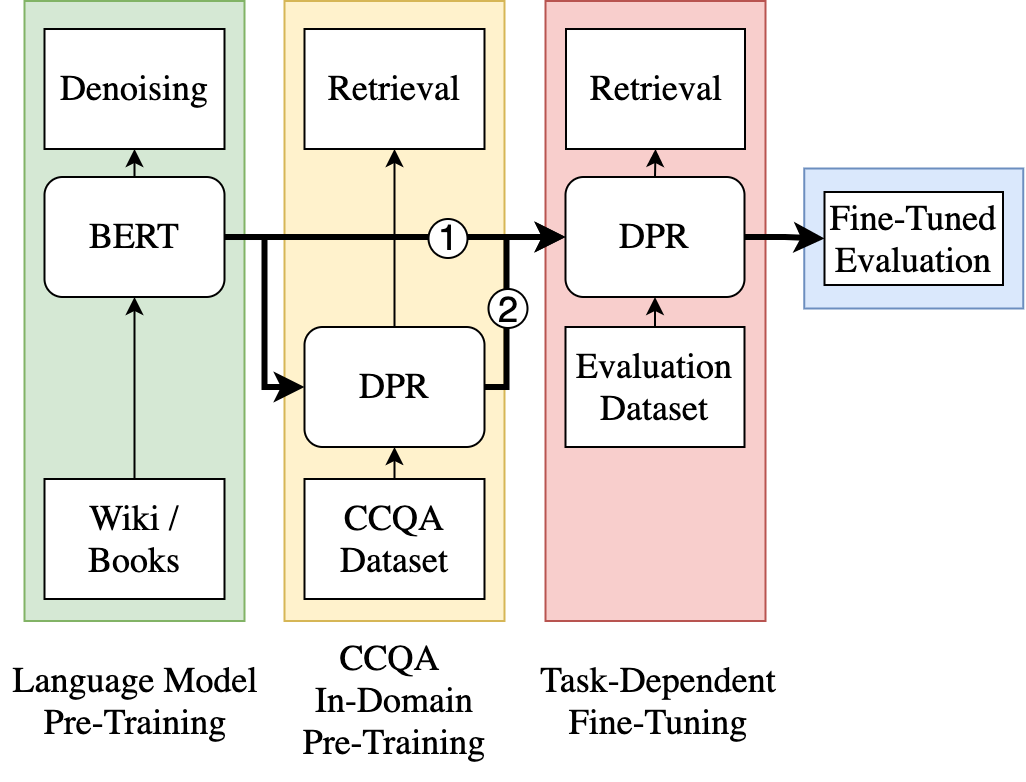}
    \caption{High-level overview of the CCQA passage retrieval in-domain pre-training step (yellow) as part of the training pipeline for DPR. Language model pre-training shown in green. Task-dependent fine-tuning presented in red. Evaluation in blue. (1) Baseline pre-training/fine-tuning pipeline, (2) In-domain pre-training/fine-tuning pipeline.}
    \label{fig:high_level_retrieval}
    \vspace{-3mm}
\end{figure}

\subsection{Passage Retrieval} 
\label{pass_ref}
\subsubsection{Task}
For the passage retrieval task, an important component of most open-book QA systems (e.g., \citet{NEURIPS2020_6b493230, izacard2021leveraging}), models aim to extract a set of evidence passages from a large collection of documents through conditional ranking. 
To align our corpus with the passage retrieval task, we aggregate every question into a single data point, consisting of the question itself, alongside all available answers as either positive or negative contexts. If available, answer votes are used as a proxy to determine positive and negative (sometimes called ``hard-negative") contexts. Following the practice in \citet{fan2019eli5}, we assign every answer with at least 2 more upvotes than downvotes as a positive context and all other answer as negative. If answer votes are not available, we use the accepted/suggested label (shown in Figure~\ref{fig:format}) as an indicator for positive and negative contexts. In the absence of either criterion, we use all available answers as positive contexts. 

\subsubsection{Models \& Training}
For passage retrieval, we choose the Dense Passage Retriever (DPR) \cite{karpukhin2020dense}, used in a variety of popular end-to-end open-book QA models, such as RAG \citep{NEURIPS2020_6b493230} and FiD \citep{izacard2021leveraging}. As shown in Figure~\ref{fig:high_level_retrieval}, we start with the vanilla DPR model based on BERT \cite{devlin-etal-2019-bert} and in-domain pre-train using questions and positive/negative passages from the CCQA dataset (yellow box in Figure~\ref{fig:high_level_retrieval}), similar to \citet{ouguz2021domain}. In line with the training-flows of the closed-book models, we train DPR using either the vanilla setup (pre-training $\rightarrow$ fine-tuning) or the in-domain pre-training approach (pre-training $\rightarrow$ in-domain pre-training $\rightarrow$ fine-tuning), shown as circles (1) and (2) in Figure~\ref{fig:high_level_retrieval}, respectively.

\subsubsection{Datasets \& Metrics}
Following the original DPR paper \citep{karpukhin2020dense}, we evaluate the passage retrieval task on the NQ-Short and TQA datasets presented in section \ref{datasets}, using the top-20 and top-100 retrieval accuracy (Acc@20/Acc@100) measures.

\begin{table}[t]
\centering
\setlength{\tabcolsep}{0.18em}
\scalebox{0.9}{
\begin{tabular}{lrrrr}
\toprule
\multirow{2}{*}{Model} & \multicolumn{2}{c}{TQA} & \multicolumn{2}{c}{NQ-Short} \\
 & \multicolumn{1}{c}{Acc@20} & \multicolumn{1}{c}{Acc@100} & \multicolumn{1}{c}{Acc@20} & \multicolumn{1}{c}{Acc@100} \\
\midrule
DPR & 79.4 & 85.0 & 78.4 & 85.4 \\
DPR v2 & 79.5 & 85.3 & 78.3 & 85.6 \\
CCQA DPR & \textbf{80.0} & \textbf{85.6} & \textbf{79.1} & \textbf{86.3} \\
\bottomrule
\end{tabular}}
\caption{Fine-tuned Dense Passage Retriever (DPR) accuracy measure on the TQA and NQ-Short datasets. DPR represents the original DPR model \cite{karpukhin2020dense}, DPR v2 \citep{ouguz2021domain} indicates the updated codebase. CCQA DPR uses our CCQA pre-trained DPR model for retrieval fine-tuning.}
\label{tab:ob_results}
\end{table}

\begin{table}
    \centering
    \setlength{\tabcolsep}{0.28em}
    \scalebox{0.9}{
    \begin{tabular}{lrrrrr}
        \toprule
        Bench. (test) & TQA & NQ-S & NQ-L & ELI5 & GooAQ \\
        \midrule
        Bench. (train) & 11.9 & 4.9 & 5.2 & 3.0 & 26.9 \\
        CCQA (train) & 0.4 & 1.9 & 2.3 & 0.5 & 26.9 \\
        \bottomrule
    \end{tabular}}
    \caption{8-gram question overlap (in \%) between training sets and benchmark test-sets (inspired by \citet{radford2019language}). \textit{Bench (train)} refers to the overlap between the respective training- and test-portion of the benchmark datasets, \textit{CCQA (train)} identified overlaps between our dataset and the test-splits. False positive rate upper-bound by $\frac{1}{10^8}$. All inputs are normalized and lower-cased. NQ-S=\allowbreak NQ-Short, NQ-L=\allowbreak NQ-Long.}
    \label{tab:bloom}
\vspace{-3mm}
\end{table}

\subsubsection{Hyper-Parameters}
We use the default DPR hyper-parameters whenever possible \citep{karpukhin2020dense}. For in-domain pre-training, we limit training to $800k$ steps using a batch-size of $1{,}536$ samples. During fine-tuning, we restrict the number of updates to $20k$ steps with a batch-size of $128$.
The best checkpoint is selected based on the Mean Reciprocal Rank (MRR) measure, following \citet{ouguz2021domain}. We do not perform any hyper-parameter search.

\subsubsection{Results} 
\label{open_book}
For the passage retrieval experiments, we compare our CCQA in-domain pre-trained DPR model against the vanilla DPR model published in \citet{karpukhin2020dense}, as well as the recently enhanced version \citep{ouguz2021domain}. Table \ref{tab:ob_results} contains our empirical results, showing consistent improvements of our CCQA DPR model over the vanilla baselines. More specifically, the in-domain CCQA pre-training step increases the top-20 and top-100 accuracy score on the TQA benchmark dataset by over half a point, while the performance gap on NQ-Short shows consistent improvement of over $0.7\%$.

\subsection{Evaluation Fairness: Dataset Overlap}
\label{overlap}
With modern pre-training approaches using increasingly large datasets, accidental overlaps between pre-training corpora and benchmark datasets become more and more common \citep{lewis2021question}. To analyze this threat to the integrity of our dataset and empirical analysis, we follow \citet{radford2019language} and evaluate the 8-gram question overlap of our CCQA training portion with the test-split of benchmark datasets using bloom filters. Table \ref{tab:bloom} shows a consistently smaller question overlap between CCQA and the benchmark test set, compared to the benchmark training split itself.

\section{Conclusion and Future Work} 
In this work, we presented our new web-scale CCQA dataset for in-domain model pre-training. Orthogonal to recent efforts on improving task-specific pre-training objectives, we show our dataset generation process, followed by detailed insights into key corpus dimensions of this new, large-scale, natural, and diverse question-answering dataset. In a set of empirical evaluations, we confirm the initial intuition that the corpus presents a valuable resource for open-domain question-answering research. In our zero-shot, low-resource and fine-tuned experiments for open- and closed-book QA tasks, we show promising results across multiple model architectures. 
With around $130$ million question-answer pairs ($55$M unique) as well as additional meta-data, our CCQA dataset presents a versatile source of information, which has a large variety of applications in future work (e.g., question summarization, answer rating, answer ranking and many more). 

\section{Ethical Considerations}
We now discuss the three major ethical considerations impacting this paper:

\paragraph{Hate-speech, Harmful Gender and Racial Biases:} 
With general web-data potentially containing hate-speech and harmful gender and racial biases, we believe that our extracted dataset based on the schema.org annotations is less impacted by these issues, with the schema.org annotation representing a good proxy for high-quality, professionally curated websites. As a result, we believe that the severity of this issue is significantly reduced. Furthermore, in our human evaluation, we find no signs of the above mentioned biases. We leave computational approaches to determine dataset biases for future work (e.g., the Word Embedding Association Test \citep{Caliskan2017SemanticsDA} and Sentence Encoder Association Test \citep{may2019measuring}).

\paragraph{Data Availability:}
We do not directly provide the CCQA dataset, but enable third parties to generate the corpus through our published dataset generation scripts available at \url{https://github.com/facebookresearch/CCQA}.

\paragraph{Hallucinations and Factual Errors:}
As shown in the evaluation section, our model is able to generate reasonable answers for factoid and long-form questions. The inferred answers are fluent and human-like, but may also contain hallucinations and factual errors, especially for the challenging closed-book question-answering task. Without a guarantee of the predicted answers being factually correct, they can potentially spread misinformation if not properly corrected.

\subsubsection*{Acknowledgments}
We want to thank the anonymous reviewers for their valuable feedback and acknowledge our colleagues and peers for their help throughout this project, specifically Anchit Gupta, Gagan Aneja and Patrick Lewis for their valuable input on important decisions.

\bibliography{anthology,custom}
\bibliographystyle{acl_natbib}

\clearpage
\appendix
\onecolumn

\section{CCQA Dataset Generation Algorithm}
\label{app:algo}
\begin{algorithm}[H]
\caption{CCQA Dataset Generation Procedure}
\label{alg:algo_1}
\onecolumn
\begin{algorithmic}
\For{document $\in$ CommonCrawl}
\If{"schema.org/Question" in document} \Comment{Webpage contains schema.org annotation}
      \State tree $\leftarrow$ parse\_html(document)
      \State questions $\leftarrow$ find\_question\_root(tree)
      \For{question\_sub\_tree in questions}
      \State question\_sub\_tree $\leftarrow$ clean\_question\_subtree(question\_sub\_tree)
      \EndFor
    \State questions $\leftarrow$ convert\_to\_json(questions)
    \State save(questions)
\Else{}
    \State skip document
\EndIf
\EndFor \\
\Procedure{find\_question\_root}{node} \Comment{Pre-order traversal, return when question found}
  \If{node.itemtype == "https://schema.org/Question"}
  \State return node
  \EndIf
  \For{child in node.children()}
  \State node $\leftarrow$ find\_question\_root(child)
  \State nodes.append(node)
  \EndFor
  \State return nodes
\EndProcedure \\

\Procedure{clean\_question\_subtree}{node} \Comment{Post-order traversal, clean elements bottom-up}
    \For{child in node}
    \State child $\leftarrow$ clean\_question\_subtree(child)
    \EndFor
    \If{"itemtype" $|$ "itemprop" in node.attributes()}
        \For{attribute in node.attributes()}
            \If{not attribute.starts\_with("item" $|$ "content" $|$ "date") }
                \State attribute.remove()
            \EndIf
        \EndFor
    \Else
        \State replace\_node\_with\_children(node)
    \EndIf
\EndProcedure
\end{algorithmic}
\end{algorithm}

\newpage

\section{Detailed Topic Distribution}
\label{app:topic_dist}
\begin{table*}[h]
    \centering
    \scalebox{0.8}{
    \begin{tabular}{llllll}
        \toprule\addlinespace
        Topic & \multicolumn{5}{c}{Top 5 Appearances in CCQA} \\
        \addlinespace\midrule\addlinespace
        Top-Level & Regional (38.90\%) & Society (21.14\%) & Business (8.36\%) & Sports (7.04\%) & Rec. (6.20\%) \\
        \addlinespace\midrule\addlinespace
        Regional & \makecell[l]{North America\\(61.48\%)}  & \makecell[l]{Europe \\(34.69\%)} & \makecell[l]{Asia\\(1.28\%)} & & \\
        \addlinespace\midrule\addlinespace
        Society & \makecell[l]{Issues \\(76.89\%)} & \makecell[l]{Religion \\(18.39\%)} & \makecell[l]{Philosophy \\(2.36\%)} & \makecell[l]{Law \\(1.41\%)} & \\
        \addlinespace\midrule\addlinespace
        Business & \makecell[l]{Industrial Goods \\(13.41\%)}  & \makecell[l]{Energy \\(9.75\%)}  & \makecell[l]{Textiles \\(9.75\%)}  & \makecell[l]{Construction \\(7.31\%)}  & \makecell[l]{Business Services \\(6.09\%)} \\
        \addlinespace\midrule\addlinespace
        Sports & \makecell[l]{Golf \\(81.08)}  & \makecell[l]{Aquatiques \\(10.81\%)}  & \makecell[l]{Events \\(2.70\%)}  & \makecell[l]{Water Sports \\(2.70\%)}  & \makecell[l]{Lacrosse \\(1.35\%)} \\
        \addlinespace\midrule\addlinespace
        Recreational & \makecell[l]{Food \\(56.92)}  & \makecell[l]{Outdoors \\(23.07\%)}  & \makecell[l]{Travel \\(12.30\%)}  & \makecell[l]{Motorcycles \\(3.07\%)}  & \makecell[l]{Pets \\(1.53\%)} \\
        \addlinespace\bottomrule
    \end{tabular}}
    \caption{Fine-grained CCQA dataset topic distribution of $1000$ randomly chosen domains retrieved through the DMOZ/Curlie annotation at https://curlie.org/. Only showing sub-topics with $\geq$ 1\%.}
    \label{tab:data_distribution_detailed}
\end{table*}

\section{Detailed Question Word Distribution}
\label{app:q_dist}
\begin{table*}[h]
    \centering
    \scalebox{0.85}{
    \begin{tabular}{lllllllll}
        \toprule\addlinespace
        Question-Word & What & How & When & Which & Where & Why & Who & Whose \\
        \addlinespace\midrule\addlinespace
        Frequency & \makecell[l]{5.3M \\ (36.20\%)} & \makecell[l]{4.3M \\ (29.80\%)} & \makecell[l]{1.4M \\ (9.68\%)} &  \makecell[l]{1.4M \\ (9.64\%)} & \makecell[l]{881k \\ (6.04\%)} & \makecell[l]{717k \\ (4.92\%)} & \makecell[l]{514k \\ (3.53\%)}  & \makecell[l]{25k \\ (0.17\%)} \\
        \addlinespace\bottomrule
    \end{tabular}}
    \caption{Question word distribution for all 8 English question words with their number of appearance in the CCQA corpus and their relative frequency.}
    \label{tab:q_words_detailed}
\end{table*}

\newpage

\section{HTML Markup Tag Distribution}
\label{app:html_dist}
\begin{table*}[h]
    \centering
    \scalebox{0.95}{
    \begin{tabular}{llllll}
        \toprule\addlinespace
        Rank & \multicolumn{5}{c}{HTML Markup Tag Distribution} \\
        \addlinespace\midrule\addlinespace
        1-5 & p (28.48\%) & a (14.89\%) & br (14.87\%) & li (10.04\%) & span (5.77\%) \\
        \addlinespace\midrule\addlinespace
        6-10 & strong (4.93\%) & code (4.59\%) & em (2.79) & div (2.38\%) & ul (2.27\%) \\
        \addlinespace\midrule\addlinespace
        11-15 & pre (1.80\%) & b (1.70\%) & blockquote (1.14\%) & h3 (0.89\%) & td (0.88\%) \\
        \addlinespace\midrule\addlinespace
        16-20 & h2 (0.48\%) & ol (0.42\%) & tr (0.42\%) & h1 (0.35\%) & i (0.24\%) \\
        \addlinespace\midrule\addlinespace
        21-25 & sup (0.17\%) & tbody (0.12\%) & table (0.12\%) & u (0.12\%) & sub (0.11\%) \\
        \addlinespace\bottomrule
    \end{tabular}
    }
    \caption{Distribution of the 25 most common HTML tags in CCQA.}
    \label{tab:data_markup_detailed}
\end{table*}

\section{Sensibility and Answerability Examples}
\label{app:sens_ans}
\begin{table*}[h]
    \centering
    \begin{tabular}{llll}
        \toprule\addlinespace
        Metric & Type & Example & Explanation \\
        \addlinespace\midrule
        Q-sensibility & Pos & What languages do you speak? & \makecell[l]{Q-Sensible, since question \\ internally makes sense} \\
        & Neg & How blue is the number 7? & \makecell[l]{Not Q-Sensible, since  question \\ internally makes no sense} \\
        Q-answerability & Pos & \makecell[l]{How can I purchase affordable \\ Flats in Vancouver?} & \makecell[l]{Q-Answerable, since a \\ single answer exists} \\
        & Neg & What languages do you speak? & \makecell[l]{Not Q-Answerable, since no single \\ answer exists, but depends on \\ the (unavailable) context} \\
        QA-sensibility & Pos & \makecell[l]{Which is the busiest month \\ to travel from London to Delhi? \\ $\rightarrow$ July} & \makecell[l]{QA-Sensible, since question and \\ answer make sense together} \\
        & Neg & \makecell[l]{How can I purchase affordable \\ Flats in Vancouver? \\ $\rightarrow$ There are many affordable \\ Flats available.} & \makecell[l]{Not QA-Sensible, since answer \\ does not answer the question} \\
        \bottomrule
    \end{tabular}
    \caption{Examples and explanations for Question-sensibility (Q-sensibility), Question-answerability (Q-answerability) and QA-sensibility. Pos = Positive example, Neg = Negative example.}
    \label{tab:sens_ans_detailed}
\end{table*}

\newpage

\section{Full Set of Low Resource Experiments}
\label{app_full_res}
\begin{figure*}[h]
    \centering
    \begin{tikzpicture}
        \begin{axis}[
                width=0.5*\textwidth,
                height=6.5cm, 
                legend pos=south east,
                ymajorgrids=true,
                xmajorgrids=true,
                bar width=1.5mm,
                ymin=6, ymax=29,
                y label style={at={(axis description cs:-0.18, .5)},anchor=north},
                ylabel={\bf NQ-Long},
                symbolic x coords={$0$, $10^1$, $10^2$, $10^3$, $10^4$, $10^5$, full},
            ]
            
            \addplot coordinates {
                ($0$, 9.16) ($10^1$, 15.06) ($10^2$, 15.4) ($10^3$, 16.12) ($10^4$, 17.32) ($10^5$, 21.49) (full, 22.09)
            };
            \addlegendentry{Vanilla}
            \node at (axis cs:$0$,7.16){\small\textcolor{blue}{9.16}};
            \node at (axis cs:$10^1$,13.06){\small\textcolor{blue}{15.06}};
            \node at (axis cs:$10^2$,13.4){\small\textcolor{blue}{15.4}};
            \node at (axis cs:$10^3$,14.12){\small\textcolor{blue}{16.12}};
            \node at (axis cs:$10^4$,15.32){\small\textcolor{blue}{17.32}};
            \node at (axis cs:$10^5$,19.49){\small\textcolor{blue}{21.49}};
            \node at (axis cs:full,20.09){\small\textcolor{blue}{22.09}};
            
            \addplot coordinates {
                 ($0$, 13.63) ($10^1$, 15.31) ($10^2$, 16.06) ($10^3$, 16.47) ($10^4$, 16.76) ($10^5$, 21.39) (full, 22.05)
            };
            \addlegendentry{CCQA}
            \node at (axis cs:$0$,15.63){\small\textcolor{red}{13.63}};
            \node at (axis cs:$10^1$,17.31){\small\textcolor{red}{15.31}};
            \node at (axis cs:$10^2$,18.06){\small\textcolor{red}{16.06}};
            \node at (axis cs:$10^3$,18.47){\small\textcolor{red}{16.47}};
            \node at (axis cs:$10^4$,18.76){\small\textcolor{red}{16.76}};
            \node at (axis cs:$10^5$,23.39){\small\textcolor{red}{21.39}};
            \node at (axis cs:full,24.05){\small\textcolor{red}{22.05}};
            
        \end{axis}
    \end{tikzpicture}
    \begin{tikzpicture}
        \begin{axis}[
                width=0.5*\textwidth,
                height=6.5cm, 
                legend pos=south east,
                ymajorgrids=true,
                xmajorgrids=true,
                bar width=1.5mm,
                ymin=6, ymax=29,
                symbolic x coords={$0$, $10^1$, $10^2$, $10^3$, $10^4$, $10^5$, full},
            ]
            \addplot coordinates {
                ($0$, 14.39) ($10^1$, 15.3) ($10^2$, 15.74) ($10^3$, 16.08) ($10^4$, 18.48) ($10^5$, 24.49) (full, 25.36)
            };
            \addlegendentry{Vanilla}
            \node at (axis cs:$0$,12.59){\small\textcolor{blue}{14.39}};
            \node at (axis cs:$10^1$,13.3){\small\textcolor{blue}{15.3}};
            \node at (axis cs:$10^2$,13.74){\small\textcolor{blue}{15.74}};
            \node at (axis cs:$10^3$,14.08){\small\textcolor{blue}{16.08}};
            \node at (axis cs:$10^4$,16.48){\small\textcolor{blue}{18.48}};
            \node at (axis cs:$10^5$,22.49){\small\textcolor{blue}{24.49}};
            \node at (axis cs:full,23.36){\small\textcolor{blue}{25.36}};
            
            \addplot coordinates {
                 ($0$, 15.08) ($10^1$, 15.63) ($10^2$, 15.89) ($10^3$, 15.9) ($10^4$, 17.91) ($10^5$, 24.08) (full, 24.73)
            };
            \addlegendentry{CCQA}
            \node at (axis cs:$0$,17.08){\small\textcolor{red}{15.08}};
            \node at (axis cs:$10^1$,17.63){\small\textcolor{red}{15.63}};
            \node at (axis cs:$10^2$,17.89){\small\textcolor{red}{15.89}};
            \node at (axis cs:$10^3$,17.90){\small\textcolor{red}{15.90}};
            \node at (axis cs:$10^4$,19.91){\small\textcolor{red}{17.91}};
            \node at (axis cs:$10^5$,26.08){\small\textcolor{red}{24.08}};
            \node at (axis cs:full,26.73){\small\textcolor{red}{24.73}};
            
        \end{axis}
    \end{tikzpicture}
    
   \begin{tikzpicture}
        \begin{axis}[
                width=0.5*\textwidth,
                height=6.5cm, 
                legend pos=south east,
                ymajorgrids=true,
                xmajorgrids=true,
                bar width=1.5mm,
                ymin=5, ymax=33,
                y label style={at={(axis description cs:-0.18, .5)},anchor=north},
                ylabel={\bf GooAQ},
                symbolic x coords={$0$, $10^1$, $10^2$, $10^3$, $10^4$, $10^5$, full},
            ]
            
            \addplot coordinates {
                ($0$, 8.92) ($10^1$, 16.07) ($10^2$, 16.63) ($10^3$, 18.13) ($10^4$, 19.15) ($10^5$, 24.77) (full, 24.7)
            };
            \addlegendentry{Vanilla}
            \node at (axis cs:$0$,6.92){\small\textcolor{blue}{8.92}};
            \node at (axis cs:$10^1$,14.07){\small\textcolor{blue}{16.07}};
            \node at (axis cs:$10^2$,14.63){\small\textcolor{blue}{16.63}};
            \node at (axis cs:$10^3$,16.13){\small\textcolor{blue}{18.13}};
            \node at (axis cs:$10^4$,17.15){\small\textcolor{blue}{19.15}};
            \node at (axis cs:$10^5$,22.77){\small\textcolor{blue}{24.77}};
            \node at (axis cs:full,22.7){\small\textcolor{blue}{24.7}};
            
            \addplot coordinates {
                 ($0$, 15.46) ($10^1$, 17.12) ($10^2$, 17.33) ($10^3$, 18.44) ($10^4$, 19.11) ($10^5$, 24.83) (full, 25.35)
            };
            \addlegendentry{CCQA}
            \node at (axis cs:$0$,17.46){\small\textcolor{red}{15.46}};
            \node at (axis cs:$10^1$,19.12){\small\textcolor{red}{17.12}};
            \node at (axis cs:$10^2$,19.33){\small\textcolor{red}{17.33}};
            \node at (axis cs:$10^3$,20.44){\small\textcolor{red}{18.44}};
            \node at (axis cs:$10^4$,21.11){\small\textcolor{red}{19.11}};
            \node at (axis cs:$10^5$,26.83){\small\textcolor{red}{24.83}};
            \node at (axis cs:full,27.35){\small\textcolor{red}{25.35}};
        \end{axis}
    \end{tikzpicture}
    \begin{tikzpicture}
        \begin{axis}[
                width=0.5*\textwidth,
                height=6.5cm, 
                legend pos=south east,
                ymajorgrids=true,
                xmajorgrids=true,
                bar width=1.5mm,
                ymin=5, ymax=33,
                symbolic x coords={$0$, $10^1$, $10^2$, $10^3$, $10^4$, $10^5$, full},
            ]
            \addplot coordinates {
               ($0$, 14.99) ($10^1$, 16.51) ($10^2$, 16.91) ($10^3$, 17.48) ($10^4$, 19.68) ($10^5$, 27.19) (full, 29.36)
            };
            \addlegendentry{Vanilla}
            \node at (axis cs:$0$,12.99){\small\textcolor{blue}{14.99}};
            \node at (axis cs:$10^1$,14.51){\small\textcolor{blue}{16.51}};
            \node at (axis cs:$10^2$,14.91){\small\textcolor{blue}{16.91}};
            \node at (axis cs:$10^3$,15.48){\small\textcolor{blue}{17.48}};
            \node at (axis cs:$10^4$,17.68){\small\textcolor{blue}{19.68}};
            \node at (axis cs:$10^5$,25.19){\small\textcolor{blue}{27.19}};
            \node at (axis cs:full,27.36){\small\textcolor{blue}{29.36}};
            
            \addplot coordinates {
                 ($0$, 15.85) ($10^1$, 17.12) ($10^2$, 17.63) ($10^3$, 17.77) ($10^4$, 19.64) ($10^5$, 27.19) (full, 29.09)
            };
            \addlegendentry{CCQA}
            \node at (axis cs:$0$,17.85){\small\textcolor{red}{15.85}};
            \node at (axis cs:$10^1$,19.12){\small\textcolor{red}{17.12}};
            \node at (axis cs:$10^2$,19.63){\small\textcolor{red}{17.63}};
            \node at (axis cs:$10^3$,19.77){\small\textcolor{red}{17.77}};
            \node at (axis cs:$10^4$,21.64){\small\textcolor{red}{19.64}};
            \node at (axis cs:$10^5$,29.19){\small\textcolor{red}{27.19}};
            \node at (axis cs:full,31.09){\small\textcolor{red}{29.09}};
            
        \end{axis}
    \end{tikzpicture}
    
   \begin{tikzpicture}
        \begin{axis}[
                width=0.5*\textwidth,
                height=6.5cm, 
                legend pos=south east,
                ymajorgrids=true,
                xmajorgrids=true,
                bar width=1.5mm,
                ymin=6, ymax=18.5,
                y label style={at={(axis description cs:-0.18, .5)},anchor=north},
                ylabel={\bf ELI5},
                x label style={at={(axis description cs:0.5,-0.13)},anchor=north},
                xlabel={\bf T5 Small},
                symbolic x coords={$0$, $10^1$, $10^2$, $10^3$, $10^4$, $10^5$, full},
            ]
            
            \addplot coordinates {
                ($0$, 7.55) ($10^1$, 13.99) ($10^2$, 15.16) ($10^3$, 15.55) ($10^4$, 16.11) ($10^5$, 16.21) (full, 16.28)
            };
            \addlegendentry{Vanilla}
            \node at (axis cs:$0$,6.8){\small\textcolor{blue}{7.55}};
            \node at (axis cs:$10^1$,13.1){\small\textcolor{blue}{13.99}};
            \node at (axis cs:$10^2$,14.2){\small\textcolor{blue}{15.16}};
            \node at (axis cs:$10^3$,14.6){\small\textcolor{blue}{15.55}};
            \node at (axis cs:$10^4$,15.1){\small\textcolor{blue}{16.11}};
            \node at (axis cs:$10^5$,15.2){\small\textcolor{blue}{16.21}};
            \node at (axis cs:full,15.3){\small\textcolor{blue}{16.28}};

            \addplot coordinates {
                 ($0$, 15.28) ($10^1$, 15.5) ($10^2$, 15.80) ($10^3$, 15.94) ($10^4$, 16.12) ($10^5$, 16.24) (full, 16.33)

            };
            \addlegendentry{CCQA}
            \node at (axis cs:$0$,16){\small\textcolor{red}{15.28}};
            \node at (axis cs:$10^1$,16.25){\small\textcolor{red}{15.50}};
            \node at (axis cs:$10^2$,16.5){\small\textcolor{red}{15.80}};
            \node at (axis cs:$10^3$,16.75){\small\textcolor{red}{15.94}};
            \node at (axis cs:$10^4$,16.9){\small\textcolor{red}{16.12}};
            \node at (axis cs:$10^5$,17){\small\textcolor{red}{16.24}};
            \node at (axis cs:full,17.1){\small\textcolor{red}{16.33}};
            
    \end{axis}
    \end{tikzpicture}
    \begin{tikzpicture}
        \begin{axis}[
                width=0.50*\textwidth,
                height=6.5cm, 
                legend pos=south east,
                ymajorgrids=true,
                xmajorgrids=true,
                bar width=1.5mm,
                ymin=6, ymax=18.5,
                x label style={at={(axis description cs:0.5,-0.13)},anchor=north},
                xlabel={\bf T5 Base},
                symbolic x coords={$0$, $10^1$, $10^2$, $10^3$, $10^4$, $10^5$, full},
            ]
            \addplot coordinates {
               ($0$, 12.27) ($10^1$, 14.69) ($10^2$, 15.38) ($10^3$, 15.42) ($10^4$, 15.53) ($10^5$, 15.76) (full, 16.58)
            };
            \addlegendentry{Vanilla}
            \node at (axis cs:$0$,11.5){\small\textcolor{blue}{12.27}};
            \node at (axis cs:$10^1$,14){\small\textcolor{blue}{14.69}};
            \node at (axis cs:$10^2$,14.5){\small\textcolor{blue}{15.38}};
            \node at (axis cs:$10^3$,14.6){\small\textcolor{blue}{15.42}};
            \node at (axis cs:$10^4$,14.75){\small\textcolor{blue}{15.53}};
            \node at (axis cs:$10^5$,15){\small\textcolor{blue}{15.76}};
            \node at (axis cs:full,15.75){\small\textcolor{blue}{16.58}};
            
            \addplot coordinates {
                 ($0$, 15.69) ($10^1$, 15.8) ($10^2$, 15.91) ($10^3$, 16.02) ($10^4$, 16.13) ($10^5$, 16.48) (full, 16.64)
            };
            \addlegendentry{CCQA}
            \node at (axis cs:$0$,16.5){\small\textcolor{red}{15.69}};
            \node at (axis cs:$10^1$,16.6){\small\textcolor{red}{15.80}};
            \node at (axis cs:$10^2$,16.7){\small\textcolor{red}{15.91}};
            \node at (axis cs:$10^3$,16.8){\small\textcolor{red}{16.02}};
            \node at (axis cs:$10^4$,16.9){\small\textcolor{red}{16.13}};
            \node at (axis cs:$10^5$,17.3){\small\textcolor{red}{16.48}};
            \node at (axis cs:full,17.4){\small\textcolor{red}{16.64}};
            
        \end{axis}
    \end{tikzpicture}
    
    \caption{Low resource experiments comparing the Rouge-L score of vanilla T5 Small (left) and T5 Base (right) with our CCQA pre-trained models on NQ-long (top), GooAQ (center) and ELI5 (bottom).}
    \label{fig:app_low_resource}
    \vspace{-3mm}
\end{figure*}
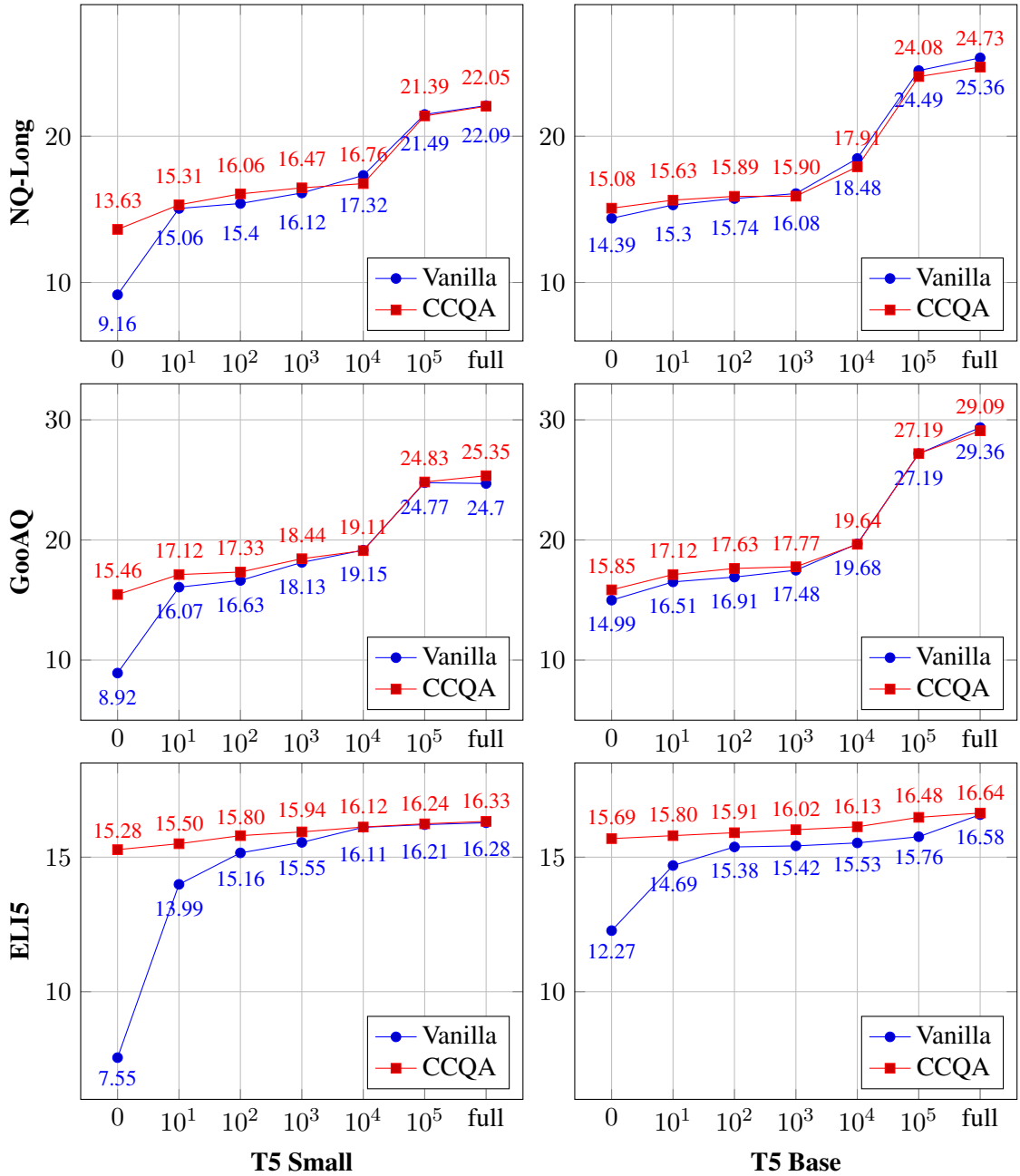

\newpage

\section{Qualitative Dataset Examples}
\label{quali}
\begin{spverbatim}
{
"Language":"-",
"Fasttext_language":"en",
"URI":"https://www.geograph.ie/faq3.php?q=multiple+account",
"UUID":"a5e97da2-f688-42af-8626-73a38fa8d06f",
"WARC_ID":"CC-MAIN-20201026031408-20201026061408-00221",
"Questions":[
      {
         "name_markup":"Can I change my name to a <b>pseudonym</b> on a submission ?",
         "Answers":[
            {
               "text_markup":"You can submit all your photos under a pseudonym by changing the name on your Profile<span><a>http://www.geograph.org.uk/profile.php</a></span>(link top write on most pages). Note that by doing this, the name will be changed on all photos you have previously submitted from the account. These may already have been used elsewhere, crediting the name originally shown. <br> You can change the credit on an individual image, for instance if you asked someone else to take it for you, but the name on your profile will still be shown on the photo page and the photographer name will still link back to your profile. <br> You can open another account under a pseudonym but this will need to be done from a different email address and you will have to take care which account you are signed in with before submitting, making changes or posting in the forums.",
               "status":"acceptedAnswer"
            }
         ]
      }
   ]
}
\end{spverbatim}
\newpage
\begin{spverbatim}
{
"Language":"en-US",
"Fasttext_language":"en",
"URI":"https://www.catholicfaithstore.com/Store/Products/SKU/b0d/
       St-Olgas-Cross-Medal.html",
"UUID":"94def557-e521-493a-babd-b63c5e030e62",
"WARC_ID":"CC-MAIN-20210308174330-20210308204330-00337",
"Questions":[
      {
         "name_markup":"How do I care for my sterling silver?",
         "Answers":[
            {
               "text_markup":"<p>Sterling Silver Cleaning Instructions</p><ul><li>NEVER use a sterling silver cleaning solution on your jewelry. It will take off the protective coating.</li><li>Take a half cup of warm water and a few drops of mild dishwashing liquid soap and mix together.</li><li>With a soft clean cotton cloth&#160;dip the cloth into the soapy water getting it moist.</li><li>Use the moist cloth to wipe the surface of your sterling silver jewelry.</li><li>Take the just cleaned jewelry and run under clear water for a few seconds to&#160;wash away any soap.</li><li>Allow jewelry to dry before storing</li></ul><p>Other things to remember: When not wearing your sterling silver jewelry, keep it in an air-tight container or zip lock bag. Avoid household clean products getting in contact with the jewelry. And take off your jewelry when you swim, shower or are washing dishes.</p><p>For a more detailed explanation see<a>5 Easy-To-Follow Steps for Cleaning Your Sterling Silver Jewelry</a></p>",
               "status":"acceptedAnswer"
            }
         ]
      }
   ]
}

\end{spverbatim}
\newpage
\begin{spverbatim}
{
"Language":"-",
"Fasttext_language":"en",
"URI":"https://quant.stackexchange.com/questions/39510/
       software-for-american-basket-option-pricing-using-longstaff
       -schwartz-least-squar",
"UUID":"e059deaf-3d73-4517-88a0-8abb8ad74972",
"WARC_ID":"CC-MAIN-20210305183324-20210305213324-00585",
"Questions":[
      {
         "author":"Bananach",
         "name_markup":"<a>Software for American basket option pricing using Longstaff-Schwartz/Least Squares Monte Carlo method</a>",
         "text_markup":"<p>Is there free software (preferably in Python) that computes American basket (high-dimensional!) option prices in the Black Scholes model using the Longstaff-Schwartz algorithm (also known as Least Squares Monte Carlo)?</p>~<p>Optimally, I want to be able to control the number of basis functions, the number of Monte Carlo samples and the number of time steps used.</p>",
         "date_created":"2018-04-30T09:16:33",
         "upvote_count":"1",
         "answer_count":"1",
         "Answers":[
            {
               "author":"byouness",
               "text_markup":"<p>QuantLib is what you are looking for. It is free/open source library written in C++, it is available in Python as well (via SWIG):<a>https://www.quantlib.org/install/windows-python.shtml </a></p>~<p>Examples are shipped with QuantLib and among them some show how to price options.</p><p>To get a feel for what it's like, you can check this blog post, explaining how to price an American option on a single asset using a binomial tree in Python:~<a>http://gouthamanbalaraman.com/blog/ american-option-pricing-quantlib-python.html</a></p>",
               "status":"acceptedAnswer",
               "upvote_count":"1",
               "comment_count":"1"
            }
         ]
      }
   ]
}
\end{spverbatim}
\newpage
\begin{spverbatim}
{
"Language":"en",
"Fasttext_language":"en",
"URI":"https://wwwmybizpro.invoicera.com/expense-management.html",
"UUID":"8cfe986c-4f33-4a2a-98f1-32aab3811533",
"WARC_ID":"CC-MAIN-20210512100748-20210512130748-00544",
"Questions":[
      {
         "name_markup":"Do I need any new IT infrastructure to get the best use out of this software?",
         "Answers":[
            {
               "text_markup":"NO! Invoicera simply integrates with your current ERP and CRM. It comes with the simplest self-explanatory user-interface for you to use. You do not need any extra guidance with your Invoicera.",
               "status":"acceptedAnswer"
            }
         ]
      }
   ]
}
\end{spverbatim}
\end{document}